%% file: main.tex
\documentclass{article}

\usepackage[preprint]{neurips_2026}

\usepackage[utf8]{inputenc}
\usepackage[T1]{fontenc}
\usepackage{amsmath}
\usepackage{amssymb}
\usepackage{amsthm}
\usepackage{booktabs}
\usepackage{graphicx}
\usepackage{microtype}
\usepackage{xcolor}
\usepackage{hyperref}
\usepackage{url}

\hypersetup{
  pdftitle={Improving Math Reasoning through Value-guided Informative Search},
  pdfauthor={Shaohuai Liu, Yuning Wu, Haoran Liu, Chaoqun Jia, Devin Chen, Kai Wei}
}

\newcommand{\added}[2][]{#2}
\newcommand{\deleted}[2][]{}
\newcommand{\replaced}[3][]{#2}

\newtheorem{theorem}{Theorem}
\newcommand{\method}{APIVIS}

\title{Improving Math Reasoning through Value-guided Informative Search}

\author{
\textbf{Shaohuai Liu}\textsuperscript{1,2,*} \quad
\textbf{Yuning Wu}\textsuperscript{1,*} \quad
\textbf{Haoran Liu}\textsuperscript{1}
\\[2pt]
\textbf{Chaoqun Jia}\textsuperscript{1} \quad
\textbf{Devin Chen}\textsuperscript{1} \quad
\textbf{Kai Wei}\textsuperscript{1,\textdagger}
\\[4pt]
\textsuperscript{1}Amazon
\\
\textsuperscript{2}Department of Electrical and Computer Engineering,
Texas A\&M University
\\[2pt]
\texttt{\{shaohuai,yuningwu,liuhr,enzojia,devichen,kaiwe\}@amazon.com}
\\
\texttt{liushaohuai5@tamu.edu}
\\[2pt]
\textsuperscript{*}Equal contribution.
\quad
\textsuperscript{\textdagger}Corresponding author.
}

\begin{document}

\maketitle

\input{sections/abstract}

\input{sections/introduction}
\input{sections/method}
\input{sections/experiments}
\input{sections/discussion}

\label{sec:content-end}
\clearpage
\bibliographystyle{plainnat}
\bibliography{references}

\clearpage
\appendix
\input{sections/related_work}
\input{sections/chunk_definition}
\input{sections/hyperparameters}
\input{sections/training_dynamics}
\input{sections/theory_appendix}
\input{sections/ablation_table}

\end{document}

%% file: sections/abstract.tex
\begin{abstract}
Reinforcement learning with verifiable rewards (RLVR) has substantially
improved the mathematical reasoning capabilities of large language models.
Recent work introduces search into RLVR rollouts to increase trajectory
diversity, but diversity alone does not ensure that the search-induced rollout
policy improves upon the current policy. To address this gap, we propose
\method, a training-time framework that adapts finite-budget Gumbel search to
chunk-level mathematical reasoning. \method{} combines direct and searched
responses within each rollout group, allowing improvements found by search to
produce informative relative rewards. It further applies selective supervision
to search-improved tokens, preserving a learning signal when uniform group
rewards render GRPO ineffective. We show that exact value-guided selection
improves the expected verifier reward at each searched state and that this
guarantee extends to the complete rollout policy, with a corresponding
approximate guarantee under bounded value-estimation error.
Experiments on widely recognized mathematical reasoning benchmarks
\added[id=CR]{and different model scales} demonstrate substantial improvements
over competitive search-based methods, validating the effectiveness of
\method{}.
\end{abstract}

%% file: sections/introduction.tex
\section{Introduction}

Reinforcement learning with verifiable rewards (RLVR) improves mathematical
reasoning by optimizing groups of responses scored by an exact answer checker
\citep{shao2024deepseekmath,deepseekr1}. While GRPO and its relatives dominate
current RLVR pipelines, their learning signal is largely determined by rollout
diversity. Standard pipelines draw several complete responses independently
from the current policy, yet token-level randomness often produces surface
variations of the same reasoning path. This limits the reasoning strategies
available for learning; when the responses also share the same final return,
their group-relative advantages vanish altogether. RLVR therefore faces both an
exploration problem and a learnability problem: its rollout operator must
improve the trajectory distribution and expose a preference that the policy can
learn.

A natural remedy is to make trajectory generation \emph{principled} rather than
rely on independent sampling. At inference time, process-reward and value-guided
search already provide effective test-time scaling by selecting among candidate
continuations \citep{lightman2023verify,wang2025vgs}. This success has motivated
training-time search for rollout diversity or learning targets, but existing
methods improve returned responses rather than establish an improved rollout
policy (Appendix~\ref{app:related}). Greater diversity may restore relative
contrast, and extra search may improve correctness, but neither guarantees that
the search-induced policy dominates the current policy. We therefore ask:
\emph{can search act as a guaranteed policy-improvement operator inside RLVR,
and can that improvement remain learnable when group returns are identical?}

Gumbel MuZero offers an inspiration but not a direct solution. Its central idea
is to define a searched policy by monotonically reweighting the prior policy
with completed action values, ensuring that the expected value under the
searched distribution is no lower than under the prior; Gumbel action sampling
and sequential halving approximate this operator under a limited simulation
budget \citep{danihelka2022gumbel}. This construction assumes a finite,
enumerable action set whose prior probabilities and action values refer to the
same decision. In language reasoning, actions are variable-length chunks, the
candidate set is itself sampled from the policy, and values are estimated
probabilities of reaching correct answers given the response prefixes. Search
may commit several chunks within one completion.
Translating the improvement operator therefore requires defining the policy
induced jointly by candidate generation and value-based selection, proving that
the local inequality survives random candidate sets and composes across a
complete trajectory, and quantifying the effect of value error. The resulting
local preference must then be transferred to the language model even when
identical final returns remove the group-relative signal.

We introduce \method, which treats reasoning chunks as search actions and uses
value-guided Gumbel search as an improvement operator inside online RLVR. Each
rollout group mixes direct and searched responses, while a selective auxiliary
loss trains only on search-improved tokens. The former converts better search
outcomes into group-relative contrast; the latter preserves an
improvement-aligned signal when GRPO group returns are uninformative,
\added[id=CR]{provided that the value-improvement filter admits searched
tokens. In the reported configuration, successful and failed trajectories have
the same InfoSFT weight, so an all-failure group can receive InfoSFT supervision
even though its group-relative GRPO gradient is zero.} We summarize our contributions as
follows:
\begin{itemize}
  \setlength{\itemsep}{0pt}
  \setlength{\parskip}{0pt}
  \setlength{\parsep}{0pt}
  \setlength{\topsep}{2pt}
  \item We formulate value-guided selection over sampled reasoning chunks and
  prove statewise and complete-response policy improvement, with an explicit
  degradation bound under value-estimation error.
  \item We propose \method, which combines finite-budget chunk search,
  mixed-policy rollout groups, and selective supervision to
  \replaced[id=CR]{distill accepted search-improved decisions, including
  uniform-return groups for which the group-relative gradient
  vanishes.}{make searched improvements learnable even from uniform-reward
  groups.}
  \item We
  \replaced[id=CR]{evaluate \method{} at Qwen3-1.7B and Qwen3-4B scales, obtaining
  consistent improvements over direct RLVR and search-based baselines.}{evaluate
  \method{} against GRPO, LATR, and TS-LLM under a controlled training protocol
  and show substantial improvements on established mathematical reasoning
  benchmarks.}
\end{itemize}

\begin{figure}[t]
\centering
\includegraphics[width=\linewidth]{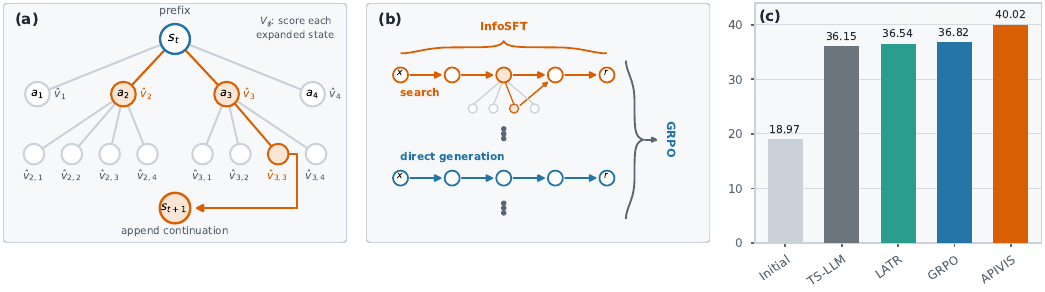}
\caption{\method{} at a glance. (a) A frozen value model scores expanded chunk
states while Gumbel sequential halving allocates the search budget and commits
the highest-$Q$ continuation. (b) Each RLVR group mixes three direct and three
searched responses; return contrast trains GRPO, while improved searched tokens
provide selective supervision. (c) Mean Avg@16 across MATH-500, AIME-2024,
AIME-2025, and OlympiadBench, where \method{} outperforms all baselines.}
\label{fig:overview}
\end{figure}

%% file: sections/method.tex
\section{Value-guided informative search}

\paragraph{Chunk-level reasoning process.}
For problem $x$, state $s$ contains the prompt and committed reasoning, and
action $a$ is the next variable-length reasoning chunk. The transition appends
the chunk, $s'=s\circ a$, and the only reward is terminal correctness
$r(x,y)\in\{0,1\}$. We write
\begin{equation}
Q^\pi(s,a)=\mathbb{E}[r\mid s,a,\text{ then follow }\pi],
\label{eq:q}
\end{equation}
where $\pi$ is the current language-model policy. Chunks keep the tree shallow
and align with partial solutions scored by token-level value models
\citep{wang2025vgs}; Appendix~\ref{app:chunks} specifies the operational chunk
boundaries.

\paragraph{Value-guided chunk search.}
At fixed chunk boundaries, $\pi$ proposes $C$ candidates and a frozen value
model scores their prefixes (Figure~\ref{fig:overview}a). Gumbel sequential
halving allocates $S$ simulations by repeatedly pruning weak roots and expanding
survivors. The rollout commits the final principal chunk and resumes decoding;
the completed response receives the usual terminal verifier reward.
\added[id=CR]{Concretely, root candidates are sampled independently from the
current policy and assigned independent Gumbel variables. Simulations are
divided across sequential-halving phases. Each simulation descends from its
assigned root through least-visited children with deterministic tie breaking,
evaluates an unexpanded or terminal leaf with the frozen value model, expands a
nonterminal leaf with policy-sampled chunks, and backs the leaf value up through
visit counts, value sums, and their mean. At each phase boundary, the
lower-scoring half of the roots is removed. After the final phase, the
highest-scoring surviving root and its highest-value principal continuation are
committed; discarded branches never enter the training batch.}

\paragraph{Mixed-policy rollout groups.}
Each six-response group contains three direct samples from $\pi$ and three from
the search-induced policy $\mu$ (Figure~\ref{fig:overview}b). GRPO normalizes
all six returns together, so improvements found by search create within-group
contrast while direct responses anchor learning to the current policy.

\paragraph{Policy-improvement view.}
The core property does not depend on a particular tree implementation. Let
$\mathcal{A}_s=(a_1,\ldots,a_C)\sim\pi(\cdot\mid s)^C$ and
$q_i=Q^\pi(s,a_i)$. For any non-decreasing transform $h$, define a
value-improved distribution over candidate slots,
\begin{equation}
\rho_h(i\mid\mathcal{A}_s)
=\frac{\exp(h(q_i))}{\sum_{j=1}^C\exp(h(q_j))}.
\label{eq:rho}
\end{equation}
This includes increasingly greedy value selection as the scale of $h$ grows.

\begin{theorem}[Selective chunk-level improvement]
\label{thm:improvement}
For every fixed candidate list,
\begin{equation}
\sum_i\rho_h(i\mid\mathcal{A}_s)q_i
\ge \frac{1}{C}\sum_i q_i.
\label{eq:candidate-improvement}
\end{equation}
Averaging over candidate lists therefore induces a chunk policy $\pi_h$ with
$\mathbb{E}_{a\sim\pi_h}Q^\pi(s,a)\ge V^\pi(s)$. If selection uses estimates
$\widehat Q$ satisfying $|\widehat Q-Q^\pi|\le\epsilon$ on all candidates, the
right-hand side becomes $V^\pi(s)-2\epsilon$. Let a completion policy $\mu$ use
the improved rule only at the scheduled searched states and use $\pi$ elsewhere.
Then exact values imply $V^\mu(s_0)\ge V^\pi(s_0)$ at the initial prompt state;
with at most $K$ searched states,
$V^\mu(s_0)\ge V^\pi(s_0)-2K\epsilon$.
\end{theorem}

The proof combines monotone-reweighting covariance with the finite-horizon
performance-difference identity (Appendix~\ref{app:proof}).
\emph{Trajectory-level improvement} does not require search at every chunk
boundary: unsearched states contribute zero advantage, so gains at searched
states compose over the complete response.
Our implementation approximates Eq.~\eqref{eq:rho} using finite-budget
sequential halving and a frozen value model. Thus the theorem identifies the
target operator and exposes value error; it is not an unconditional claim about
the learned parameterized policy.

\paragraph{Learning from informative continuations.}
All committed tokens train with the same group-relative loss as the sampling
baseline. A committed searched chunk enters mask $\mathcal{M}$ only when its
backed-up value improves on the pre-search state or the policy's direct
candidate. The auxiliary objective is
\begin{equation}
\mathcal{L}_{\rm InfoSFT}
=-\frac{1}{Z}\sum_i w_i\sum_{t\in\mathcal{M}_i}
\log\pi_\theta(y_t^{(i)}\mid x,y_{<t}^{(i)}),
\qquad
\mathcal{L}=\mathcal{L}_{\rm GRPO}+w_{\rm I}\mathcal{L}_{\rm InfoSFT}.
\label{eq:objective}
\end{equation}
\added[id=CR]{In the reported experiment, $w_i=1$ for both verified-success
and failed trajectories; terminal correctness is not an admission requirement.
If the weighted mask is empty, we define the auxiliary loss to have exactly
zero gradient.}
\replaced[id=CR]{When group returns are identical, GRPO is zero. InfoSFT
supplies supervision only when $\mathcal{M}$ is nonempty and its trajectory
weight is nonzero. Under the reported weighting this can occur for both
all-success and all-failure groups. Mixed-return groups may receive both
group-relative learning and selective supervision on admitted searched
trajectories.}{When group returns are identical, GRPO is zero but a nonempty
$\mathcal{M}$ still supplies supervision aligned with the improved search
policy. InfoSFT therefore transfers local search gains even when
relative-reward learning fails.}

%% file: sections/experiments.tex
\section{Experiments}

\subsection{Setup}
\replaced[id=CR]{We evaluate Qwen3-1.7B-Base and Qwen3-4B-Base.}{All
experiments initialize from Qwen3-4B-Base.} Every run uses the same ordered
Open-RS training data \citep{dang2025openrs}, verifier, optimizer, and data
schedule: 64 prompts per step, six completions per prompt, and 100 training
steps.
\replaced[id=CR]{At both scales we compare the initial policy, GRPO, LATR, and
\method{}; the 4B study additionally includes TS-LLM, while the 1.7B study
includes a matched MCTS-alpha + GRPO baseline.}{We compare the initial policy,
GRPO, LATR, TS-LLM, and \method{}.} LATR retains its native branching rule.
TS-LLM and the matched MCTS-alpha baseline use MCTS with the same per-search
simulation budget as the Gumbel search in \method{}. Search placement in
\method{} is fixed across trajectories. Each \method{} group contains three
directly sampled completions and three search-generated completions.
\added[id=CR]{All mathematical training results use one training seed. Search
methods additionally incur candidate-generation and value-model inference
cost.}
\added[id=CR]{Over the archived 4B runs, \method{} uses 20.06 logged training
hours (160.51 GPU-hours), compared with 6.32 hours for GRPO, 38.64 for LATR,
and 40.16 for TS-LLM; Appendix~\ref{app:compute} gives the full accounting.}

We evaluate direct policy decoding on MATH-500
\citep{hendrycks2021math}, AIME-2024, AIME-2025, and OlympiadBench
\citep{he2024olympiadbench}. For each problem, Avg@16 is the mean verifier
accuracy over 16 sampled completions at temperature $0.7$ and top-$p=0.95$.
\replaced[id=CR]{The validation artifact is a deterministic $0.5\%$ split of
Open-RS generated with seed 42 (35 prompts). Each validation point uses four
direct-policy samples per prompt. For every headline model-comparison row, we
select the saved checkpoint ranked first by validation mean@4, with earlier-step
tie breaking, and only then run the four full test benchmarks.}{For each run, we
select the checkpoint with the best validation performance and report its full
benchmark evaluation.} The reported \method{} runs use fixed 3-direct/3-search
in training rollouts; Appendix~\ref{app:hyperparameters} lists all
hyperparameters.

\subsection{Main Results}
Table~\ref{tab:main} compares the rollout and search strategies at both model
scales. With Qwen3-4B-Base, \method{} achieves $40.02$ average Avg@16,
exceeding GRPO, LATR, and TS-LLM by $3.20$, $3.48$, and $3.87$ percentage
points, respectively, and performs best on all four benchmarks. Neither LATR
($36.54$) nor TS-LLM ($36.15$) improves the average over GRPO ($36.82$).
Search-based generation alone therefore does not automatically translate into
a stronger directly decoded policy under this protocol.

Relative to GRPO, \method{} gains $2.32$ points on MATH-500, $2.71$ on
AIME-2024, $4.17$ on AIME-2025, and $3.59$ on OlympiadBench. Relative to the
LATR baseline, the gains are $1.41$, $4.17$, $3.96$, and $4.37$ points,
respectively. With Qwen3-1.7B-Base, \method{} reaches $27.79$, compared with
$25.26$ for GRPO, $22.59$ for LATR, and $25.10$ for MCTS-alpha + GRPO: gains
of $2.53$, $5.20$, and $2.69$ points, respectively.

\begin{table}[t]
\caption{\replaced[id=CR]{Validation-rank-1 Avg@16 verifier accuracy (\%) at
two model scales. Each trained row uses one training seed.}{Avg@16 verifier
accuracy (\%) with Qwen3-4B-Base.}}
\label{tab:main}
\centering
\small
\setlength{\tabcolsep}{2.8pt}
\begin{tabular}{lccccc}
\toprule
Method & MATH-500 & AIME-24 & AIME-25 & Olympiad & Avg. \\
\midrule
\textit{Qwen3-1.7B-Base} & 47.60 & 4.38 & 2.29 & 18.61 & 18.22 \\
\quad + GRPO & 63.50 & 6.25 & 3.12 & 28.16 & 25.26 \\
\quad + LATR + GRPO & 57.79 & 5.83 & 2.50 & 24.23 & 22.59 \\
\quad + MCTS-alpha + GRPO & 64.38 & 6.25 & 2.29 & 27.47 & 25.10 \\
\quad + \method{} + InfoSFT & \textbf{66.17} & \textbf{9.58} & \textbf{4.58}
& \textbf{30.80} & \textbf{27.79} \\
\midrule
\textit{Qwen3-4B-Base} & 45.44 & 4.58 & 4.17 & 21.68 & 18.97 \\
\quad + GRPO & 78.83 & 13.75 & 10.62 & 44.07 & 36.82 \\
\quad + LATR & 79.74 & 12.29 & 10.83 & 43.29 & 36.54 \\
\quad + TS-LLM & 77.39 & 13.75 & 11.04 & 42.42 & 36.15 \\
\quad + \method{} & \textbf{81.15} & \textbf{16.46} & \textbf{14.79}
& \textbf{47.66} & \textbf{40.02} \\
\bottomrule
\end{tabular}
\end{table}

\paragraph{\added[id=CR]{Training-rollout search diagnostic.}}
\added[id=CR]{Archived telemetry shows higher mean terminal reward for searched
than direct completions at both scales: a $0.0630$ lift at 1.7B and $0.0541$
at 4B. Prompt-level comparisons nevertheless include losses, so search
improves the rollout distribution on average rather than every prompt;
Figure~\ref{fig:search-reward-lift} shows the per-step dynamics, and
Appendix~\ref{app:training-diagnostics} gives the full statistics.}

\paragraph{\added[id=CR]{Value-model ranking diagnostic.}}
\added[id=CR]{At the Qwen3-1.7B step-95 checkpoint, a no-update diagnostic over
all 35 validation roots finds positive candidate-ranking signal (Spearman
$0.239$, 95\% CI $[0.035,0.433]$; top-1 lift $0.039$, 95\% CI
$[0.008,0.077]$). At the Qwen3-4B step-100 checkpoint, the rank-correlation
point estimate is positive but inconclusive (Spearman $0.146$, 95\% CI
$[-0.075,0.368]$), with zero aggregate top-1 lift ($0.000$, 95\% CI
$[-0.032,0.029]$). Monitoring such ranking statistics may provide a practical
signal that value-model quality has plateaued and that another round of value
training is warranted.}

\subsection{Ablation Analysis}
Table~\ref{tab:ablations} isolates search and selective supervision under the
same fixed 3-direct/3-search layout. Replacing MCTS-alpha with the
policy-improvement Gumbel operator raises average Avg@16 from $36.29$ to
$39.32$, a $3.03$-point gain. Adding InfoSFT gives a further $0.70$ points,
reaching $40.02$. Its largest incremental gains occur on AIME-2025 ($1.04$)
and OlympiadBench ($1.06$),
\replaced[id=CR]{consistent with, but not by itself establishing, the proposed
benefit of selective distillation when group-level contrast is absent.}{consistent
with the additional token-level signal being most useful when group-level
contrast is sparse.}
Together, these results
support both value-guided selection and explicit learning from searched
continuations.

\paragraph{\added[id=CR]{InfoSFT group diagnostic.}}
\added[id=CR]{At a fixed 1.7B step-95 checkpoint, 31 of 64 diagnostic groups
are all-zero and hence have zero group-relative GRPO signal, yet contain
10,209 positively weighted InfoSFT tokens. This verifies the effective loss
semantics, not historical group prevalence; see
Appendix~\ref{app:training-diagnostics}.}

The search-chunk budget $K$ has a non-monotonic effect
(Table~\ref{tab:k-ablation}). Increasing $K$ from $5$ to $10$ raises average
Avg@16 from $39.20$ to $40.02$, whereas $K=20$ falls slightly to $39.84$.
Although $K=20$ is strongest on AIME-2024 and OlympiadBench, it regresses on
MATH-500 and AIME-2025 and does not improve mean Pass@16 over $K=5$. Thus,
allowing more searched decisions does not automatically produce a more
learnable policy. \added[id=CR]{The corresponding mean numbers of distinct
normalized final answers per problem are $4.36$, $4.20$, and $4.37$, showing
no monotonic diversity collapse as $K$ increases; see
Appendix~\ref{app:k-diversity}.}
\added[id=CR]{An adaptive informative-state variant retains essentially the
same Avg@16 as fixed placement while reducing mean step time, although it
remains more expensive than GRPO and differs in InfoSFT weighting; see
Appendix~\ref{app:adaptive-placement}.} Appendix~\ref{app:dynamics} reports
validation dynamics.

%% file: sections/discussion.tex
\section{Discussion}
\label{sec:discussion}

\paragraph{Limitations.}
The value model is frozen throughout training. This keeps the search signal
stable and makes the comparison interpretable, but it also prevents the value
function from adapting as the policy improves; stale or miscalibrated values can
misrank chunks, and the approximate guarantee degrades with value error.
Moreover, finite-budget sequential halving only approximates the ideal improved
distribution, and our experiments consider single-turn mathematical
completions with terminal verification.
\added[id=CR]{The experiments cover two model scales but use one training seed
per configuration, and the ablations do not separately isolate the effect of
mixed direct/search group construction.}

\paragraph{Future work.}
A natural next step is to co-train the value model and policy while controlling
critic drift, so that search quality can track the evolving policy without
destabilizing the improvement operator.
\added[id=CR]{Monitoring candidate-ranking quality may provide a practical
signal that the value model has plateaued and should undergo another round of
training.} Extending \method{} to multi-turn
reasoning is also important: agentic tasks require values over dialogue and
environment states, search budgets allocated across turns, and learning from
intermediate feedback rather than a single terminal answer.

%% file: sections/related_work.tex
\section{Related Work}
\label{app:related}

\paragraph{RLVR and rollout diversity.}
GRPO normalizes verifier rewards within responses sampled for one prompt,
removing the need for a learned critic \citep{shao2024deepseekmath};
DeepSeek-R1 demonstrates the effectiveness of this recipe for mathematical
reasoning \citep{deepseekr1}. Its update, however, depends on the responses
exposing useful return differences. When ancestral samples follow similar
reasoning paths or receive the same return, group-relative learning supplies
little information. This motivates changing the rollout operator rather than
only the optimizer.

\paragraph{Search for inference-time reasoning.}
Process-reward models rank intermediate solutions and support best-of-$n$ or
tree-style test-time scaling \citep{lightman2023verify}. Value-guided search
extends this idea to partial reasoning traces scored by a token-level value
model \citep{wang2025vgs}. These methods improve the answer returned for a fixed
prompt, but they need not define a searched distribution that improves the
policy used for subsequent training.

\paragraph{Search for training-time exploration.}
LATR branches and prunes during RLVR generation to increase trajectory
diversity \citep{xing2025latr}, while ROSE resamples from uncertain positions
and assigns tree-structured credit \citep{zhao2026rose}. TS-LLM and
ReST-MCTS$^*$ instead use tree search to construct learning targets
\citep{feng2023alphazerolike,zhang2024restmcts}. These approaches
establish the utility of search during training, but do not guarantee that the
search-induced rollout policy dominates the current language-model policy.

\paragraph{Context-guided exploration and informative sampling.}
\added[id=CR]{Search is not the only way to change the rollout distribution.}
\added[id=CR]{NudgeRL
conditions generation on lightweight strategy contexts and distills the
discovered behavior back into the base policy}
\citep{lee2026nudgerl},
\added[id=CR]{while
weak-model guidance uses prefixes from a weaker policy to move rollouts outside
the current policy's typical trajectories}
\citep{shen2026weakguidance}.
\added[id=CR]{DAPO
dynamic sampling instead resamples prompts to increase the supply of groups
with nonzero relative advantages}
\citep{yu2025dapo}.
\added[id=CR]{These approaches address
closely related exploration or learnability failures without online
value-guided tree search. Our experiments do not compare them under matched
total compute.}
\added[id=CR]{Complementarily, GAUGE shows that LLM-as-a-judge evaluation of
user-simulated task-oriented agents can be poorly anchored to grounded task
success, motivating verifiable rewards when they are available}
\citep{bodhwani2026gauge}.

\paragraph{Search as policy improvement.}
Gumbel MuZero combines Gumbel action sampling and sequential halving to
approximate a value-improved policy under a small simulation budget
\citep{danihelka2022gumbel}. \method{} adapts this principle to sampled,
variable-length reasoning chunks, then couples searched continuations to both
group-relative learning and selective supervision. The distinction is that
search is treated as an explicit improvement operator, not only as a source of
diverse or high-reward trajectories.

%% file: sections/chunk_definition.tex
\section{Reasoning Chunk Definition}
\label{app:chunks}

We factor each completion into variable-length \emph{chunks} rather than
individual tokens. A state contains the prompt and all chunks committed so far;
an action is the next chunk, and the transition appends that chunk to the
response prefix. Operationally, generation of one chunk stops at the first of

\begin{itemize}
  \item a decoded blank line \verb|"\n\n"|, which marks a reasoning-step boundary;
  \item a decoded \verb|"</answer>"| tag, which closes the final answer; or
  \item the EOS token.
\end{itemize}

These boundaries are additionally subject to the remaining response budget and
a safety token limit that prevents a pathological single chunk when no boundary
fires. Truncation is applied directly to token IDs, so the retained text is
always the exact decoded prefix. This definition keeps the search tree shallow,
allows candidates to share the same prefix and KV cache, and matches the
block-wise granularity used by the frozen value model.

%% file: sections/hyperparameters.tex
\section{Core Hyperparameters}
\label{app:hyperparameters}

Table~\ref{tab:hyperparameters} summarizes the reported fixed APIVIS
configuration and the shared training and evaluation protocol.

\begin{table*}[t]
\caption{\replaced[id=CR]{Core hyperparameters shared by the Qwen3-1.7B-Base
and Qwen3-4B-Base experiments.}{Core hyperparameters for the Qwen3-4B-Base
experiments.} APIVIS settings refer to the reported fixed 3-direct/3-search
C4/S8/K10 configuration.}
\label{tab:hyperparameters}
\centering
\small
\setlength{\tabcolsep}{5pt}
\renewcommand{\arraystretch}{1.05}
\begin{tabular}{@{}p{0.14\textwidth}p{0.25\textwidth}p{0.55\textwidth}@{}}
\toprule
Category & Hyperparameter & Value \\
\midrule
Training & Base policy & \replaced[id=CR]{Qwen3-1.7B-Base or
Qwen3-4B-Base}{Qwen3-4B-Base} \\
Training & Data & \replaced[id=CR]{Open-RS, 7,000 source prompts;
deterministic seed-42 split and order}{Open-RS, 7,000 prompts, fixed order, no
shuffle} \\
Training & Seed / optimizer steps & $1$ / $100$ \\
Training & Per-step batch & 64 prompts; 6 completions per prompt (384 total) \\
Training & PPO minibatch / epochs & 16 prompts / 1 epoch \\
Optimization & Actor learning rate & $1\times10^{-6}$ \\
Optimization & Schedule & Constant; 5\% warmup (5 steps) \\
Optimization & GRPO objective & DAPO token-mean; full-completion scope; weight 1.0 \\
Optimization & KL / rollout correction & Explicit KL disabled; token-level TIS threshold 2.0, no batch normalization \\
Rollout & Training token caps & 512 prompt tokens; 3,584 completion tokens \\
Rollout & Training sampler & Temperature 1.0; top-$p$ 1.0 \\
APIVIS & Group composition & 3 direct and 3 searched completions \\
APIVIS & Search budget & $C=4$ candidates; $S=8$ simulations; $K=10$ searched chunks \\
APIVIS & Placement / commitment & Fixed, non-adaptive placement; highest-$Q$ principal path \\
APIVIS & Value model & Frozen VGS-AI/DeepSeek-VM-1.5B \\
APIVIS & InfoSFT & \replaced[id=CR]{Weight 0.1; searched-continuation scope;
failed-trajectory weight 1; terminal correctness not required}{Weight 0.1;
searched-continuation scope; failed-trajectory weight 0} \\
APIVIS & InfoSFT admission & Value gain $>0.001$ or value margin $>0.001$ \\
Evaluation & Validation / selection & \replaced[id=CR]{35 held-out Open-RS
prompts; four samples per prompt; before training and nominally every 5 steps
for search runs, every 10 steps for GRPO; select validation rank 1 by mean@4
with earliest-step tie breaking}{Before training and every 5 steps; select the
best validation checkpoint} \\
Evaluation & Full evaluation & Avg@16; 16 samples per problem; temperature 0.7; top-$p$ 0.95 \\
Evaluation & Evaluation token caps & 512 prompt tokens; 8,192 completion tokens \\
\bottomrule
\end{tabular}
\end{table*}

\subsection{Compute Accounting}
\label{app:compute}

\begin{table}[ht]
\caption{\added[id=CR]{Logged training time and approximate eight-GPU cost.
Summed time excludes downtime between resumed processes.}}
\label{tab:compute}
\centering
\small
\setlength{\tabcolsep}{5pt}
\begin{tabular}{lrrr}
\toprule
\added[id=CR]{Scale / method} & \added[id=CR]{Steps} &
\added[id=CR]{Training hours} & \added[id=CR]{GPU-hours} \\
\midrule
\added[id=CR]{4B GRPO} & \added[id=CR]{100} & \added[id=CR]{6.32} &
\added[id=CR]{50.57} \\
\added[id=CR]{4B MCTS-alpha} & \added[id=CR]{100} & \added[id=CR]{18.24} &
\added[id=CR]{145.92} \\
\added[id=CR]{4B Gumbel} & \added[id=CR]{100} & \added[id=CR]{19.19} &
\added[id=CR]{153.50} \\
\added[id=CR]{\textbf{4B \method{}}} & \added[id=CR]{\textbf{100}} &
\added[id=CR]{\textbf{20.06}} & \added[id=CR]{\textbf{160.51}} \\
\added[id=CR]{4B LATR} & \added[id=CR]{100} & \added[id=CR]{38.64} &
\added[id=CR]{309.15} \\
\added[id=CR]{4B TS-LLM} & \added[id=CR]{100} & \added[id=CR]{40.16} &
\added[id=CR]{321.31} \\
\midrule
\added[id=CR]{1.7B \method{}} & \added[id=CR]{100} & \added[id=CR]{9.96} &
\added[id=CR]{79.70} \\
\bottomrule
\end{tabular}
\end{table}

\added[id=CR]{Training hours sum the logged durations of completed optimization
steps; GPU-hours multiply this quantity by eight. Exact generated-token
counters are absent for the archived \method{} and TS-LLM runs, so
Table~\ref{tab:compute} provides transparent wall-clock accounting rather than
a token- or compute-matched efficiency comparison.}

%% file: sections/training_dynamics.tex
\section{Training Dynamics}
\label{app:dynamics}

\begin{figure}[ht]
\centering
\includegraphics[width=\linewidth]{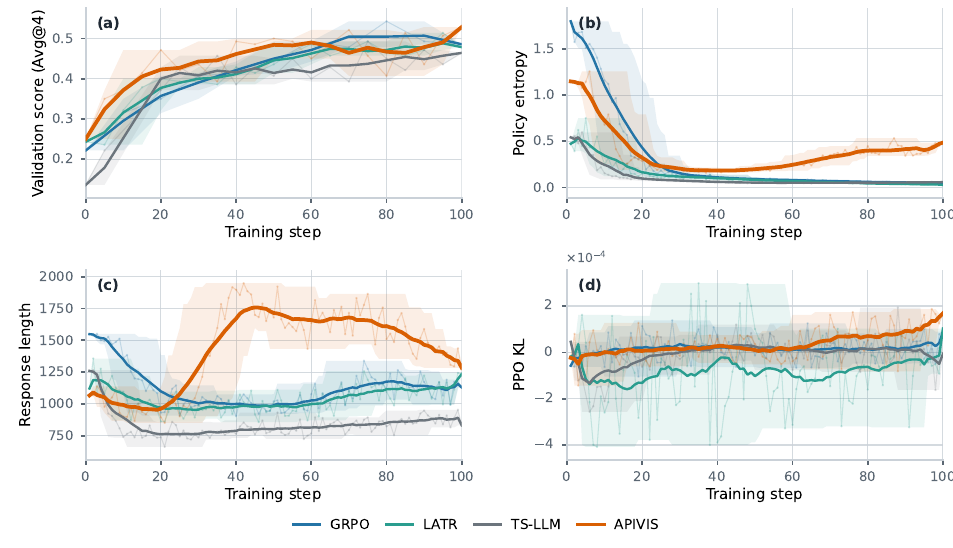}
\caption{\replaced[id=CR]{Qwen3-4B-Base training dynamics for the four methods in
Table~\ref{tab:main}: validation Avg@4, policy entropy, response length, and
PPO KL. Faint traces show logged values, solid lines show centered running
means over a 10-step radius, and bands show the local minimum--maximum range
within that window.}{Validation score (Avg@4) across training for the four
methods in Table~\ref{tab:main}.}}
\label{fig:training-dynamics}
\end{figure}

All methods improve rapidly over the first 20--40 steps, followed by substantial
validation noise. \method{} ends at $0.529$, above GRPO ($0.486$), LATR
($0.479$), and TS-LLM ($0.464$). Because the curves overlap rather than
separate uniformly, we treat the full Avg@16 evaluation in
Table~\ref{tab:main}, not any single validation point, as the primary comparison.
\added[id=CR]{The optimization panels additionally expose changes in policy
entropy, generated response length, and PPO KL rather than presenting
validation in isolation. The shaded ranges describe local temporal variation
within each single run, not uncertainty across seeds.}
\added[id=CR]{Checkpoint selection nevertheless follows validation rank 1
(maximum mean@4 with earlier-step tie breaking), before any full-test result is
used.}

%% file: sections/theory_appendix.tex
\section{Proof of Theorem~\ref{thm:improvement}}
\label{app:proof}

For a fixed candidate list, let $\rho_0(i)=1/C$ and let the random variable $q$
take value $q_i$ when $i\sim\rho_0$. Equation~\eqref{eq:rho} reweights
$\rho_0$ by $\exp(h(q))$, so
\begin{equation}
\mathbb{E}_{\rho_h}[q]
=\mathbb{E}_{\rho_0}[q]
+\frac{\operatorname{Cov}_{\rho_0}(q,\exp(h(q)))}
       {\mathbb{E}_{\rho_0}[\exp(h(q))]}
\ge \mathbb{E}_{\rho_0}[q].
\end{equation}
The covariance is nonnegative because $h$ is non-decreasing. Averaging over
candidate lists proves the exact statewise claim because a uniformly selected
candidate has marginal distribution $\pi(\cdot\mid s)$. If selection uses
$\widehat Q$, applying $|\widehat Q-Q^\pi|\le\epsilon$ before and after the
improvement inequality loses at most $2\epsilon$. Finally, the
performance-difference identity sums the statewise advantage under the
selectively searched policy. Unsearched states contribute zero, exact searched
states contribute nonnegative advantage, and at most $K$ approximate searched
states contribute no less than $-2\epsilon$ each.

%% file: sections/ablation_table.tex
\section{Controlled Search Ablations}
\label{app:ablations}

\subsection{Search Operator and Selective Supervision}

\added[id=CR]{We separate the contribution of the search operator from the
auxiliary distillation signal using three fixed 3-direct/3-search
configurations. All rows use Qwen3-4B-Base, the same data order and training
seed, full-completion GRPO, and the same $C=4$, $S=8$, $K=10$ search budget.
Replacing MCTS-alpha with Gumbel sequential halving changes only the search
operator; adding InfoSFT to the Gumbel run then tests selective supervision on
value-improving searched continuations.}

\begin{table}[ht]
\caption{Avg@16 verifier accuracy (\%) for fixed 3-direct/3-search runs.
\added[id=CR]{Each row is one training seed evaluated at its
validation-rank-1 checkpoint.}}
\label{tab:ablations}
\centering
\small
\setlength{\tabcolsep}{3.2pt}
\begin{tabular}{lccccc}
\toprule
Method & MATH-500 & AIME-24 & AIME-25 & Olympiad & Avg. \\
\midrule
MCTS-alpha + GRPO & 78.55 & 12.92 & 10.62 & 43.06 & 36.29 \\
Gumbel + GRPO & 80.70 & 16.25 & 13.75 & 46.60 & 39.32 \\
Gumbel + GRPO + InfoSFT & \textbf{81.15} & \textbf{16.46}
& \textbf{14.79} & \textbf{47.66} & \textbf{40.02} \\
\bottomrule
\end{tabular}
\end{table}

\added[id=CR]{Gumbel search improves the four-benchmark average from $36.29$
to $39.32$, a $3.03$-point gain over MCTS-alpha under the matched rollout
layout. InfoSFT contributes a further $0.70$ points, reaching $40.02$, with
the largest incremental gains on AIME-2025 and OlympiadBench. These are
controlled single-seed ablations and do not estimate training-seed
uncertainty.}

\subsection{Search-Chunk Budget}
\label{app:k-diversity}

We vary the maximum number of searched chunks $K$ while holding the fixed
3-direct/3-search layout, $C=4$ candidates, $S=8$ simulations, GRPO, and
InfoSFT fixed.

\begin{table}[ht]
\caption{Avg@16 verifier accuracy (\%) as the search-chunk budget $K$ varies.
\added[id=CR]{Each setting uses one training seed and a best-validation
checkpoint; $K=20$ uses a validation-co-best checkpoint because the strict
earliest tied checkpoint lacks a completed full evaluation.}}
\label{tab:k-ablation}
\centering
\small
\setlength{\tabcolsep}{3.2pt}
\begin{tabular}{lccccc}
\toprule
$K$ & MATH-500 & AIME-24 & AIME-25 & Olympiad & Avg. \\
\midrule
5  & 81.04 & 15.42 & 14.38 & 45.98 & 39.20 \\
10 & \textbf{81.15} & 16.46 & \textbf{14.79} & 47.66 & \textbf{40.02} \\
20 & 80.39 & \textbf{18.75} & 12.50 & \textbf{47.72} & 39.84 \\
\bottomrule
\end{tabular}
\end{table}

Mean Pass@16, the fraction of problems solved by at least one of 16 samples
averaged across benchmarks, is $62.83$, $60.28$, and $61.04$ for
$K=5,10,20$, respectively. The coverage result is also non-monotonic.
\added[id=CR]{Across the same frozen 16-sample evaluation artifacts, the mean
numbers of distinct normalized final answers per problem are $4.36$, $4.20$,
and $4.37$ for $K=5,10,20$. Distinct correct answers per solved problem are
$1.16$, $1.19$, and $1.18$, respectively. Thus, the observed performance
changes are not accompanied by a monotonic collapse in answer diversity.
Normalization extracts the final boxed answer and standardizes its string
form; it is not a symbolic-equivalence test.}

\subsection{Adaptive Informative-State Placement}
\label{app:adaptive-placement}

\added[id=CR]{The reported \method{} model uses fixed placement, but we also
evaluate adaptive placement for mathematical reasoning. At each eligible chunk
boundary, the detector first measures uncertainty under the deployed policy
sampler. Let $\ell(s)$ be the next-token logits and}
\begin{equation}
\added[id=CR]{
q_s=\operatorname{TopP}\!\left(\operatorname{softmax}(\ell(s)/\tau),p\right),
\qquad
\widetilde{\mathcal H}_{\pi}(s)
=\frac{\mathcal H(q_s)}{\log|\operatorname{supp}(q_s)|}.
}
\label{eq:adaptive-entropy}
\end{equation}
\added[id=CR]{The normalized value ambiguity of the frozen Bernoulli success
estimate is}
\begin{equation}
\added[id=CR]{
\mathcal A(s)=4V(s)\bigl(1-V(s)\bigr)\in[0,1].
}
\label{eq:adaptive-ambiguity}
\end{equation}
\added[id=CR]{The historical C4/S8 adaptive run uses the informative-state
score}
\begin{equation}
\added[id=CR]{
\mathcal I(s)=\widetilde{\mathcal H}_{\pi}(s)\mathcal A(s),\qquad
d(s)=\mathbf 1\!\left[
\widetilde{\mathcal H}_{\pi}(s)>\eta_{\mathcal H}
\ \land\
\mathcal I(s)>\kappa
\right].
}
\label{eq:adaptive-product-gate}
\end{equation}
\added[id=CR]{The entropy test is evaluated first, so states that fail it incur
no value-model detector call. Search is applied to the first $K$ accepted
boundaries. We set $\eta_{\mathcal H}=0.1$, $K=10$, and
$\kappa=0.1$ for all runs. These placement rules select
where to search; they are distinct from InfoSFT admission, which selects
searched continuations after observing value gain or margin.}

\begin{table}[ht]
\caption{\added[id=CR]{Validation-rank-1 Avg@16 and full 100-step training
telemetry for Qwen3-4B-Base. Chunks are searched chunks per designated search
trajectory; GPU-hours use eight GPUs.}}
\label{tab:adaptive-compute}
\centering
\small
\setlength{\tabcolsep}{3.2pt}
\begin{tabular}{lrrrr}
\toprule
\added[id=CR]{Method} & \added[id=CR]{Avg.} &
\added[id=CR]{Chunks} & \added[id=CR]{Step (s)} &
\added[id=CR]{GPU-h} \\
\midrule
\added[id=CR]{GRPO} & \added[id=CR]{36.82} &
\added[id=CR]{--} & \added[id=CR]{227.57} & \added[id=CR]{50.57} \\
\added[id=CR]{Adaptive C4/S8} & \added[id=CR]{39.98} &
\added[id=CR]{0.52} & \added[id=CR]{441.75} & \added[id=CR]{98.17} \\
\added[id=CR]{Adaptive C8/S16} & \added[id=CR]{40.06} &
\added[id=CR]{1.07} & \added[id=CR]{883.91} & \added[id=CR]{196.42} \\
\added[id=CR]{Fixed C4/S8} & \added[id=CR]{40.02} &
\added[id=CR]{3.91} & \added[id=CR]{722.30} & \added[id=CR]{160.51} \\
\bottomrule
\end{tabular}
\end{table}

\added[id=CR]{The C4/S8 adaptive run reduces mean searched chunks by $86.8\%$
and mean step time by $38.8\%$ relative to fixed placement, while its Avg@16
changes by only $-0.04$ points. However, it remains $1.94\times$ slower per
step than GRPO. It also uses failed-trajectory InfoSFT weight zero, whereas the
reported fixed run uses weight one. The result therefore provides a
reduced-search-compute robustness check, not a strictly compute-matched or
single-factor comparison. The wider C8/S16 adaptive variant is included for
completeness but does not provide an efficiency control.}

\subsection{Training-Time Mechanism Diagnostics}
\label{app:training-diagnostics}

\added[id=CR]{Archived training telemetry separates the three direct and three
searched completions in each \method{} group. At 1.7B, direct and searched
terminal rewards average $0.1717$ and $0.2347$, respectively, a $0.0630$ lift;
prompt-level searched-versus-direct outcomes are $22.48\%$ wins, $67.91\%$
ties, and $9.61\%$ losses. At 4B, the corresponding rewards are $0.2968$ and
$0.3509$, a $0.0541$ lift, with $25.14\%$ wins, $61.02\%$ ties, and $13.84\%$
losses.}

\begin{figure}[ht]
\centering
\includegraphics[width=\linewidth]{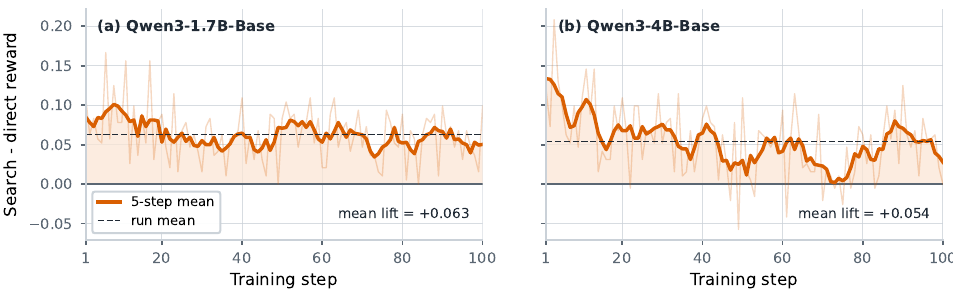}
\caption{\added[id=CR]{Per-step searched-minus-direct terminal reward for the
reported APIVIS runs. Thin lines show raw training-step values, thick lines
show centered five-step means, and dashed lines show run means. Positive values
indicate that searched completions outperform direct completions within the
same training rollout groups.}}
\label{fig:search-reward-lift}
\end{figure}

\added[id=CR]{We also run a production-faithful, no-update diagnostic on 64
six-response groups at the 1.7B step-95 checkpoint. Among the 31 all-zero
groups, for which group-relative GRPO is zero, $25.8\%$ of trajectories have a
nonempty InfoSFT mask with positive trajectory weight, totaling 10,209 admitted
tokens. The diagnostic confirms that all-failure groups can receive InfoSFT
supervision under the reported configuration. Because it is a fixed-checkpoint
diagnostic rather than archived training-wide group telemetry, it is not used
to estimate historical all-zero-group prevalence.}

%% file: references.bib
@article{shao2024deepseekmath,
  title   = {{DeepSeekMath}: Pushing the Limits of Mathematical Reasoning in Open Language Models},
  author  = {Shao, Zhihong and Wang, Peiyi and Zhu, Qihao and Xu, Runxin and Song, Junxiao and Bi, Xiao and Zhang, Haowei and Zhang, Mingchuan and Li, Y. K. and Wu, Y. and Guo, Daya},
  journal = {arXiv preprint arXiv:2402.03300},
  year    = {2024},
  doi     = {10.48550/arXiv.2402.03300},
}

@article{deepseekr1,
  title   = {{DeepSeek-R1}: Incentivizing Reasoning Capability in {LLMs} via Reinforcement Learning},
  author  = {{DeepSeek-AI}},
  journal = {arXiv preprint arXiv:2501.12948},
  year    = {2025},
  doi     = {10.48550/arXiv.2501.12948},
}

@inproceedings{danihelka2022gumbel,
  title     = {Policy Improvement by Planning with Gumbel},
  author    = {Danihelka, Ivo and Guez, Arthur and Schrittwieser, Julian and Silver, David},
  booktitle = {International Conference on Learning Representations},
  year      = {2022},
  url       = {https://openreview.net/forum?id=bERaNdoegnO}
}

@inproceedings{wang2025vgs,
  title     = {Value-Guided Search for Efficient Chain-of-Thought Reasoning},
  author    = {Wang, Kaiwen and Zhou, Jin Peng and Chang, Jonathan Daniel and Gao, Zhaolin and Kallus, Nathan and Brantley, Kiant\'e and Sun, Wen},
  booktitle = {Advances in Neural Information Processing Systems},
  year      = {2025},
  url       = {https://openreview.net/forum?id=jOsuKwiCL0}
}

@inproceedings{lightman2023verify,
  title     = {Let's Verify Step by Step},
  author    = {Lightman, Hunter and Kosaraju, Vineet and Burda, Yura and Edwards, Harri and Baker, Bowen and Lee, Teddy and Leike, Jan and Schulman, John and Sutskever, Ilya and Cobbe, Karl},
  booktitle = {International Conference on Learning Representations},
  year      = {2024},
  url       = {https://openreview.net/forum?id=v8L0pN6EOi}
}

@inproceedings{feng2023alphazerolike,
  title     = {{AlphaZero}-Like Tree-Search can Guide Large Language Model Decoding and Training},
  author    = {Wan, Ziyu and Feng, Xidong and Wen, Muning and McAleer, Stephen Marcus and Wen, Ying and Zhang, Weinan and Wang, Jun},
  booktitle = {Proceedings of the 41st International Conference on Machine Learning},
  pages     = {49890--49920},
  volume    = {235},
  series    = {Proceedings of Machine Learning Research},
  publisher = {PMLR},
  year      = {2024},
  url       = {https://proceedings.mlr.press/v235/wan24c.html}
}

@inproceedings{zhang2024restmcts,
  title     = {{ReST-MCTS*}: {LLM} Self-Training via Process Reward Guided Tree Search},
  author    = {Zhang, Dan and Zhoubian, Sining and Hu, Ziniu and Yue, Yisong and Dong, Yuxiao and Tang, Jie},
  booktitle = {Advances in Neural Information Processing Systems},
  volume    = {37},
  year      = {2024},
  doi       = {10.52202/079017-2066},
}

@article{xing2025latr,
  title   = {Lookahead Tree-Based Rollouts for Enhanced Trajectory-Level Exploration in Reinforcement Learning with Verifiable Rewards},
  author  = {Xing, Shangyu and Wang, Siyuan and Yang, Chenyuan and Dai, Xinyu and Ren, Xiang},
  journal = {arXiv preprint arXiv:2510.24302},
  year    = {2025},
  doi     = {10.48550/arXiv.2510.24302},
}

@inproceedings{zhao2026rose,
  title     = {Reinforced Efficient Reasoning via Semantically Diverse Exploration},
  author    = {Zhao, Ziqi and Ren, Zhaochun and Zou, Jiahong and Yang, Liu and Xu, Zhiwei and Ge, Xuri and Chen, Zhumin and Ma, Xinyu and Shi, Daiting and Wang, Shuaiqiang and Yin, Dawei and Xin, Xin},
  booktitle = {Proceedings of the 64th Annual Meeting of the Association for Computational Linguistics (Volume 1: Long Papers)},
  pages     = {47994--48007},
  publisher = {Association for Computational Linguistics},
  year      = {2026},
  doi       = {10.18653/v1/2026.acl-long.2216},
}

@article{dang2025openrs,
  title   = {Reinforcement Learning for Reasoning in Small {LLMs}: What Works and What Doesn't},
  author  = {Dang, Quy-Anh and Ngo, Chris},
  journal = {arXiv preprint arXiv:2503.16219},
  year    = {2025},
  doi     = {10.48550/arXiv.2503.16219},
}

@article{yu2025dapo,
  title   = {{DAPO}: An Open-Source {LLM} Reinforcement Learning System at Scale},
  author  = {Yu, Qiying and Zhang, Zheng and Zhu, Ruofei and Yuan, Yufeng and Yue, Yu and Fan, Tiantian and Liu, Gaohong and Liu, Lingjun and Liu, Xin and others},
  journal = {arXiv preprint arXiv:2503.14476},
  year    = {2025},
  url     = {https://arxiv.org/abs/2503.14476},
}

@article{lee2026nudgerl,
  title   = {Nudging Beyond the Comfort Zone: Efficient Strategy-Guided Exploration for {RLVR}},
  author  = {Lee, Chanuk and Park, Sangwoo and Kang, Minki and Hwang, Sung Ju},
  journal = {arXiv preprint arXiv:2605.15726},
  year    = {2026},
  url     = {https://arxiv.org/abs/2605.15726},
}

@article{shen2026weakguidance,
  title   = {Boosting {LLM} Exploration via Weak-Model Guidance in {RLVR}},
  author  = {Shen, Xingyu and Zhang, Huishuai and Li, Peng and Wang, Yinchun and Zhao, Dongyan},
  journal = {arXiv preprint arXiv:2608.27420},
  year    = {2026},
  url     = {https://arxiv.org/abs/2608.27420},
}

@article{bodhwani2026gauge,
  title   = {{GAUGE}: When Not to Trust {LLM}-as-a-Judge in User-Simulated Evaluation of Task-Oriented Agents},
  author  = {Bodhwani, Umesh and Tran, Thanh and Wei, Kai},
  journal = {arXiv preprint arXiv:2609.12191},
  year    = {2026},
  doi     = {10.48550/arXiv.2609.12191},
}

@inproceedings{hendrycks2021math,
  title     = {Measuring Mathematical Problem Solving with the {MATH} Dataset},
  author    = {Hendrycks, Dan and Burns, Collin and Kadavath, Saurav and Arora, Akul and Basart, Steven and Tang, Eric and Song, Dawn and Steinhardt, Jacob},
  booktitle = {NeurIPS Datasets and Benchmarks Track},
  year      = {2021},
  url       = {https://openreview.net/forum?id=7Bywt2mQsCe}
}

@inproceedings{he2024olympiadbench,
  title     = {{OlympiadBench}: A Challenging Benchmark for Promoting {AGI} with Olympiad-Level Bilingual Multimodal Scientific Problems},
  author    = {He, Chaoqun and Luo, Renjie and Bai, Yuzhuo and Hu, Shengding and Thai, Zhen and Shen, Junhao and Hu, Jinyi and Han, Xu and Huang, Yujie and Zhang, Yuxiang and Liu, Jie and Qi, Lei and Liu, Zhiyuan and Sun, Maosong},
  booktitle = {Proceedings of the 62nd Annual Meeting of the Association for Computational Linguistics (Volume 1: Long Papers)},
  pages     = {3828--3850},
  publisher = {Association for Computational Linguistics},
  year      = {2024},
  doi       = {10.18653/v1/2024.acl-long.211},
}
